\documentclass{article}

\usepackage[preprint]{corl_2026}

\usepackage{times}
\usepackage[numbers]{natbib}
\usepackage{multicol}

\usepackage{amsmath,amssymb,graphicx}
\usepackage{enumitem}

\newcommand{\AlgName}{\textit{\textbf{ForceBand}}}
\usepackage{tabularx}
\usepackage{makecell}
\usepackage{float}
\usepackage{caption}
\usepackage{wrapfig}
\usepackage{bm}
\usepackage{booktabs}
\usepackage{makecell}
\usepackage{graphicx}

\newcommand{\secv}{\vspace{-0.6em}}
\newcommand{\secvv}{\vspace{-0.3em}}

\newcommand{\paragraphc}[1]{\vspace{-0.1em}\noindent\textbf{#1}}

\begin{document}

\title{
\resizebox{\linewidth}{!}{ForceBand: Learning Forceful Manipulation with sEMG}
}

\author{
  \textbf{Botao He}$^{1,2}$ \quad
  \textbf{Zhi (Leo) Wang}$^{1, 2}$ \quad
  \textbf{Linna Kuang}$^{3}$ \quad
  \textbf{Ishaan Ghosh}$^{2}$ \\
  \textbf{Jitendra Malik}$^{1}$ \quad
  \textbf{Cornelia Fermuller}$^{2}$ \quad
  \textbf{Tingfan Wu}$^{1}$ \quad
  \textbf{Jiayuan Mao}$^{1}$ \\
  \textbf{Ruoshi Liu}$^{1,\dagger}$ \quad
  \textbf{Haozhi Qi}$^{1,\dagger}$ \quad
  \textbf{Yiannis Aloimonos}$^{2,\dagger}$ \\[3pt]
  {\normalfont $^{1}$Amazon FAR \quad
  $^{2}$University of Maryland \quad
  $^{3}$Johns Hopkins University
  } \\[2pt]
  {\normalfont $^{\dagger}$Equal advising} \quad
  {\normalfont \url{https://forceband-emg.github.io/}}
}

\makeatletter
\g@addto@macro\@maketitle{
\begin{figure}[H]
  \setlength{\linewidth}{\textwidth}
  \setlength{\hsize}{\textwidth}
    \centering
    \vspace{-2em}
    \includegraphics[width=\textwidth]{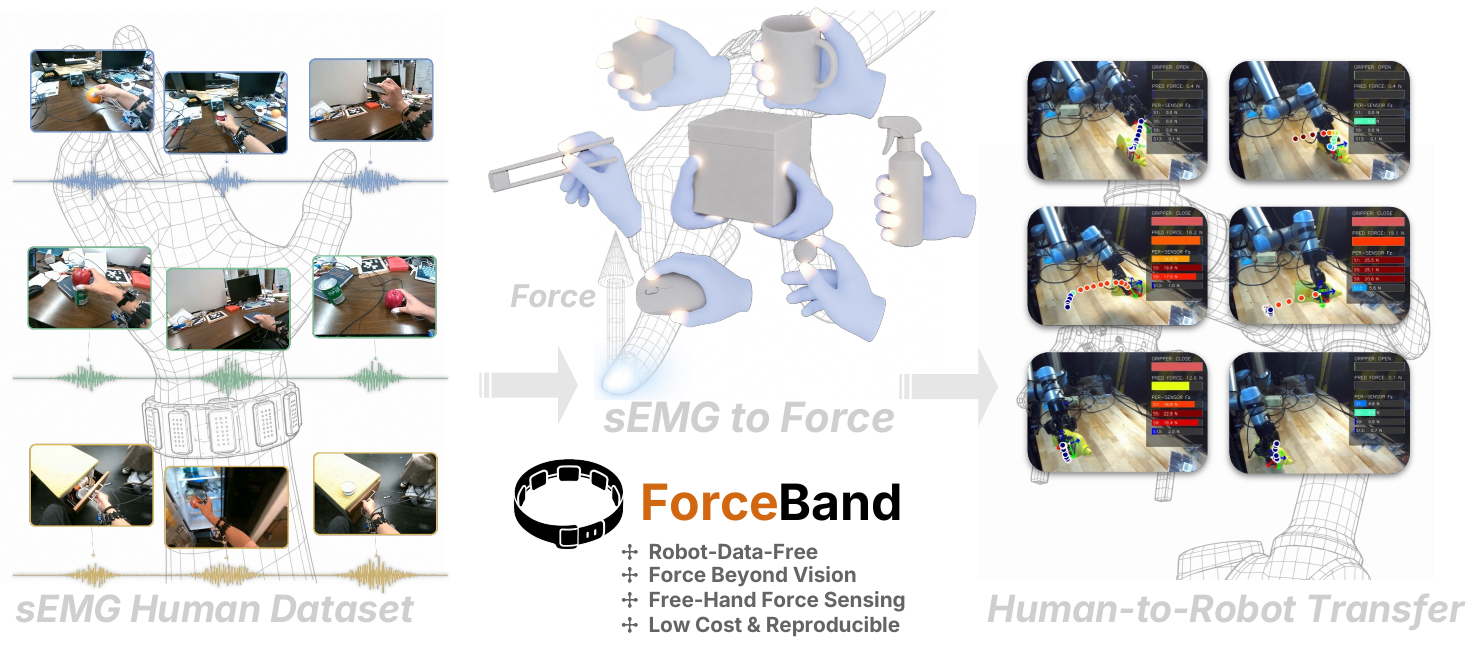}
    \captionsetup{font={footnotesize},labelfont=bf}
    \caption{\textbf{\AlgName{} learns force-aware robot policies from human demonstrations with wrist sEMG.} A human performs natural manipulation while wearing a muscle-aware surface electromyography (sEMG) wristband \textbf{(left)}. The EMG2Force model converts muscle signals into per-finger force traces \textbf{(middle)}, which are synchronized with human video to create force-enriched demonstrations. These demonstrations are retargeted to robot embodiments and used to train a force-aware policy that predicts both action and force trajectories for downstream manipulation \textbf{(right)}.}
    \vspace{-5mm}
    \label{fig:teaser}
    \end{figure}
}
\makeatother

\maketitle

\begin{abstract}
Human demonstrations are a scalable data source for learning robot manipulation policies. However, common sources of human demonstration data, such as motion-capture trajectories and internet videos, capture mostly motion and appearance, while missing the contact forces that are critical for force-sensitive manipulation.
In this paper, we introduce \AlgName{}, a low-cost wrist-worn sEMG system that turns human muscle activity into force-enriched demonstrations. We first collect a 10-hour multimodal dataset containing egocentric video, sEMG, IMU, and fingertip force measurements across diverse actions and objects.
Using this dataset, we pre-train an EMG2Force model that predicts per-finger forces from sEMG and IMU signals. After a short user-specific calibration, users can collect target-task demonstrations using only \AlgName{} and video; EMG2Force then labels these demonstrations with per-finger force traces, producing force-augmented demonstrations for robot policy learning.
Experiments show that \AlgName{} recovers fine-grained fingertip interactions with over 50\% lower force prediction error than vision-based baselines, and achieves an 87\% success rate on pick, squeeze, and place tasks that require object-specific force control across objects with diverse shapes, sizes, and weights. Project website: \url{https://forceband-emg.github.io}
\end{abstract}

\secv
\section{Introduction}
\secv

Humans achieve dexterous manipulation by continuously processing sensory signals and commanding muscle activations to produce precise finger motions and contact forces. A central objective of robot learning is to recreate this capability by learning from human data \citep{whirl,pointpolicy,p3po,hop,humanego,aina,wang2025chainofmodality}. However, these demonstrations primarily capture the visible and kinematic aspects of manipulation, while missing the contact forces that are critical for forceful manipulation.

Existing approaches to acquiring contact information face important limitations.
Vision-based methods can infer hand-object contact from video~\citep{narasimhaswamy2020detecting,yagi2021hand,hampali2020honnotate,brahmbhatt2020contactpose,tse2022s2contact,touchanything}, but the underlying forces remain highly ambiguous from appearance alone. Direct force sensing, such as tactile gloves, can provide more reliable force supervision \citep{opentouch,osmo,feeltheforce}, but these sensors can occlude visual cues and make large-scale daily demonstration collection cumbersome. These limitations motivate a sensing modality that captures force-relevant information without instrumenting the fingertips during target-task demonstrations.

Surface electromyography (sEMG) offers an unobtrusive proxy for the muscle activations that generate finger forces \citep{sun2018one,xiao2025wrist2finger,zhao2026dexemg}. However, raw sEMG signals are noisy and their relation to fingertip force varies across users and electrode placement. We address this problem in two stages. First, we train EMG2Force on a multimodal dataset containing paired wrist sEMG, IMU signals, and fingertip force measurements. Second, for each user, we perform a short calibration with fingertip force sensors to adapt the model to that user's muscle activation patterns and sensor placement. After calibration, users can collect target-task demonstrations using only \AlgName{} and video; EMG2Force labels these demonstrations with per-finger force traces, and we retarget the resulting force-enriched demonstrations to robot embodiments for force-aware policy learning.

More specifically, \AlgName{} consists of four parts:
\begin{figure}[t]
	\centering
	\includegraphics[width=\textwidth]{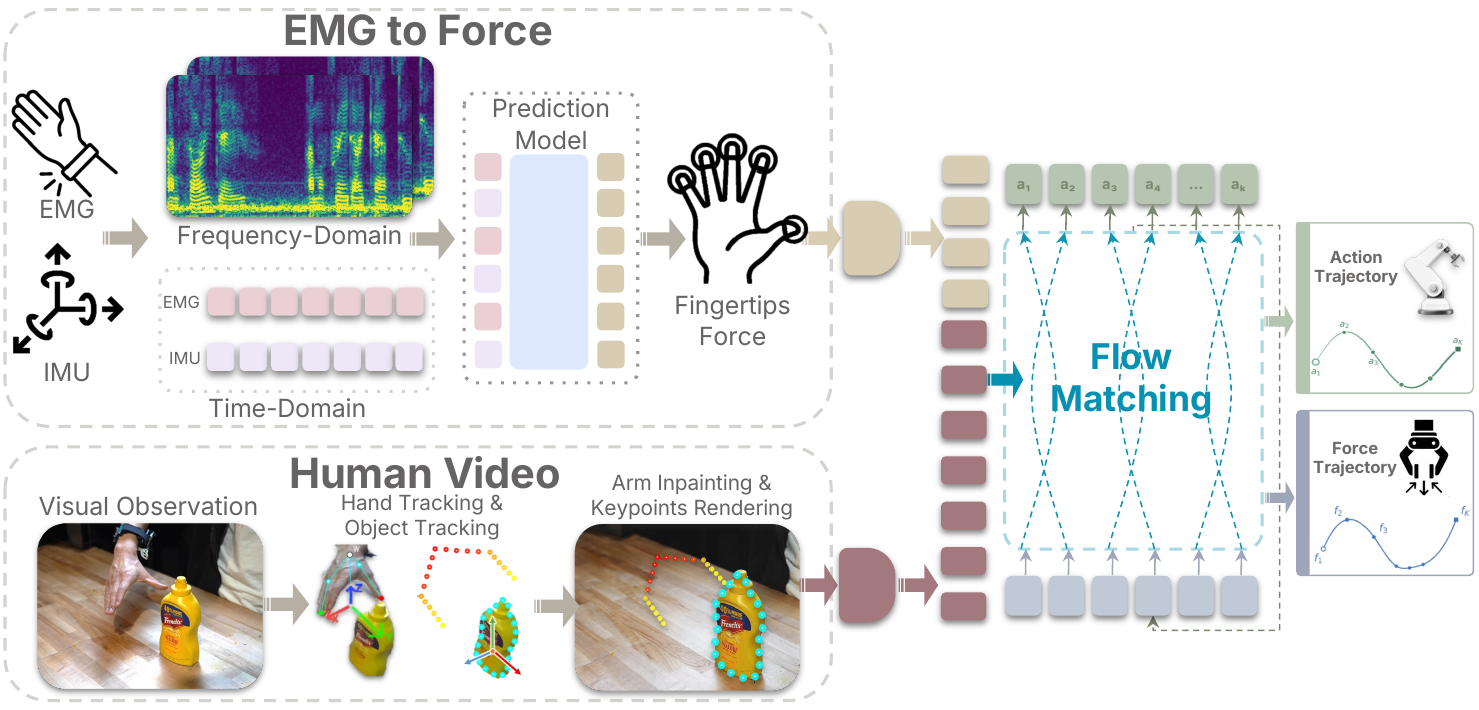}
    \caption{\textbf{\AlgName{} system architecture.} Our method predicts per-finger force traces from wrist sEMG and IMU signals by combining time domain and frequency domain representations. In parallel, human videos are transformed into robot-compatible observations. A flow matching policy is then trained to predict both action and force trajectories for forceful robot manipulation.}
	\label{fig:overview}
	\vspace{-0.5cm}
\end{figure}

\begin{itemize}[itemsep=2pt, topsep=0pt, leftmargin=15pt, parsep=0pt]
\item \textbf{Muscle-aware sEMG wristband.}
A low-cost and reproducible wrist-worn band with anatomically guided bipolar electrode placement over specific forearm muscles, paired with an IMU.

\item \textbf{Multimodal force-sensing dataset.}
A 10-hour synchronized dataset containing egocentric video, sEMG, IMU, and per-finger fingertip forces across four action categories and four gesture types, providing the supervision needed to learn the mapping from sEMG to fingertip force.

\item \textbf{EMG2Force model.}
A spectrogram-augmented transformer fuses sEMG and IMU signals to predict per-finger force traces, enabling accurate force estimation from human demonstrations for downstream policy learning.

\item \textbf{Forceful manipulation from human demonstrations.}
A policy-learning pipeline that retargets hand-object trajectories from human videos to a robot embodiment and augments them with EMG2Force-estimated forces, as well as a flow-matching visuomotor policy for force control.
\end{itemize}

We evaluate \AlgName{} at three levels: hardware design, force estimation, and force-aware robot learning. 
First, our muscle-aware electrode placement yields better force predictions than uniform wrist placement, reducing error by about 18\%. 
Second, our EMG2Force model substantially outperforms vision-based force estimation baselines, reducing prediction error by over 50\% and achieving even larger gains for occluded fingers. 
Finally, when combined with motion retargeting, the estimated forces enable learning force-aware robot policies from human videos, outperforming motion-only and gripper-only baselines on forceful manipulation tasks.
Overall, \AlgName{} provides a scalable path toward adding force supervision to human demonstration data for robot learning.

\secv
\section{Wearable Force Sensing Design}
\secv

\paragraph{Design Objectives.}

\begin{wrapfigure}{r}{0.5\textwidth}
    \vspace{-0.5cm}
    \centering
    \includegraphics[width=0.5\textwidth]{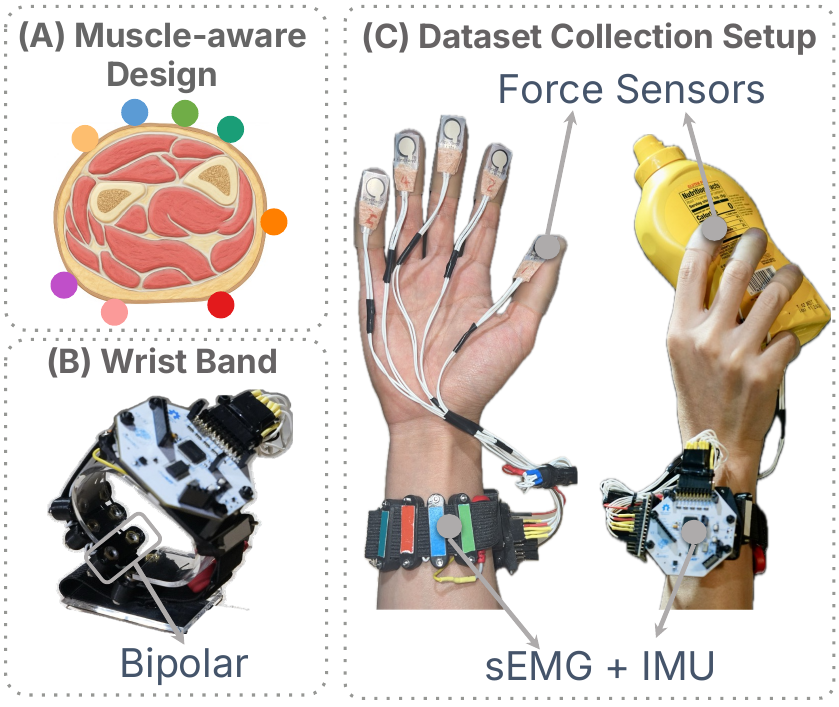}
    \vspace{-0.5cm}
    \caption{\textbf{Hardware design of \AlgName{}}. Our method combines anatomically guided wrist sEMG sensing with an IMU to capture muscle and motion signals relevant to finger-level force. Fingertip force sensors are used during \textit{dataset collection and calibration} to provide ground-truth force supervision, and are removed during target-task demonstration collection.}
    \label{fig:hardware}
    \vspace{-0.8cm}
\end{wrapfigure}

Our sEMG band is designed to accurately capture wrist muscle activity relevant to fine-grained finger control. We target a design that balances sensing accuracy, low cost, and reproducibility.
Specifically, we target two design goals.

\textbf{Accessibility: Low Cost and Easy Manufacturing.}
We aim to keep the \AlgName{} hardware low-cost and easy to reproduce by relying on commodity components and standard prototyping workflows. The mechanical structure can be fabricated with common tools such as a commercial 3D printer, while the electronics are modular and sourced from readily available parts. The total cost can be as low as \$300 depending on supplier. We open-source the complete bill of materials.

\textbf{Performance: Muscle-aware and High Precision.}
Our band collects high-quality data by combining muscle-aware electrode placement with a bipolar design. This performance is supported by three hardware choices described below: low-noise biopotential acquisition, bipolar differential sensing, and anatomically guided electrode placement.

\secvv
\subsection{Muscle-aware sEMG Wristband Design}
\secvv

We optimize the sEMG module design for recording subtle electrical potentials at the wrist through three design choices: a high-quality medical biopotential measurement chip, a bipolar electrode design, and muscle-aware electrode placement.

For the data-collection board, we adopt the OpenBCI Cyton, an open-source multi-channel biosensing board built around the ADS-1299 chip, which enables low-noise acquisition ($0.14 \mu V_{\mathrm{rms}}$ and an SNR of $119.5$).
The bipolar design, as shown in Figure \ref{fig:hardware}B, places a differential electrode pair at each site to measure the local potential difference over the target muscle region, suppressing common-mode noise and motion artifacts while preserving signal fidelity during natural manipulation. 
Finally, as shown in Figure \ref{fig:hardware}A, rather than distributing electrodes uniformly, we position them based on a \textit{non-uniform, anatomically guided pattern}, by concentrating coverage over forearm regions that provide informative signals for grasp-related finger motions, especially thumb, index, and middle finger activation. This muscle-specific placement increases the information content per channel and reduces the number of sensors required to improve wearability and efficiency without sacrificing performance, and allows the system to learn fine-grained human finger interactions and transfer the resulting force-aware representations to robot manipulation. In total, we use eight channels: seven over muscles controlling finger movement and one over a muscle controlling wrist flexion. We validate correct positioning with a short series of calibration gestures after donning the band; the full electrode sites and calibration protocol are detailed in Appendix \ref{app:placement}. The custom hardware is also flexible for modular expansion, as shown in Appendix \ref{app:hardware_extensibility}.

\secvv
\subsection{Texture-Preserving Ground-Truth Fingertip Force Sensor}
\secvv

To train and calibrate the EMG2Force model, which predicts force from sEMG signals, we use thin-film flexible force-sensitive resistors as ground-truth force sensors, as shown on the right side of Figure~\ref{fig:hardware}. Each sensor is mounted inside a transparent gel finger cot, which allows the camera to observe the natural hand appearance instead of an opaque glove surface. This design preserves the visual cues needed for hand tracking while still providing direct fingertip force measurements. All sensor wires are routed along the palm side of the hand and secured at the wrist, minimizing visual occlusion of the dorsal and lateral hand surfaces in egocentric video. \textit{After calibration, the fingertip force sensors are removed}, and fingertip forces can be estimated using \AlgName{} alone during human demonstration collection.

\secv
\section{EMG2Force: Dataset and Force Prediction Model}
\secv

\begin{figure*}[t!]
	\centering
	\includegraphics[width=\textwidth]{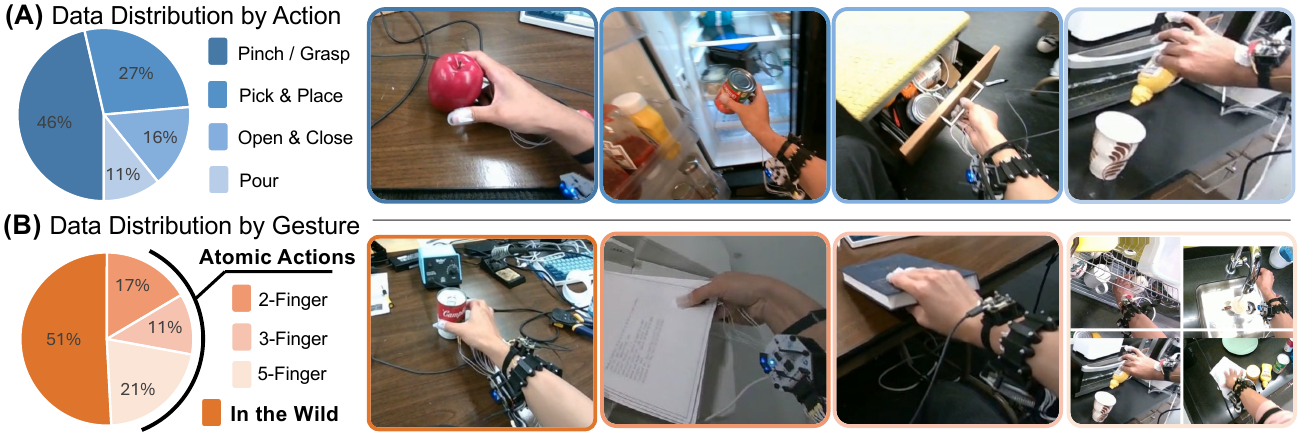}
	\caption{
		\textbf{Dataset statistics. }A dataset of synchronized egocentric video, sEMG, IMU, and per-finger force. (A) Action distribution and (B) gesture distribution, spanning atomic grasps and in-the-wild interactions with varied object shape, weight, and size. 
	}
	\label{fig:dataset}
	\vspace{-0.5cm}
\end{figure*}

\paragraphc{Multimodal force-sensing dataset.}
We collect and release a 10-hour multimodal dataset of synchronized egocentric video, sEMG, IMU, and fingertip force measurements. As shown in
Fig.~\ref{fig:dataset}A, the data spans four action categories: pinch/grasp
(46\%), pick-and-place (27\%), open/close (16\%), and pour (11\%), and four
gesture types (Fig.~\ref{fig:dataset}B): 2-finger (51\%), 3-finger (17\%), and
5-finger (11\%) atomic actions, plus 21\% in-the-wild interactions. The dataset includes objects with diverse shapes, sizes, and weights, as well as free-form daily interactions, allowing us to study force prediction beyond isolated or canonical grasps.

\paragraphc{Fingertip Force Prediction Model.} As shown in Figure \ref{fig:overview}, the model takes as input sEMG time series \(\mathbf{X}_{\mathrm{EMG}} \in \mathbb{R}^{C \times T}\) and an IMU time series \(\mathbf{X}_{\mathrm{IMU}} \in \mathbb{R}^{D \times T}\), where \(C = 8\) is the number of EMG channels, \(D = 10\) is the IMU dimension (3 acceleration, 3 gyroscope, 4 orientation), and \(T = 5s\) is the length of the time window. All sensors are sampled at \(250\,\mathrm{Hz}\), so each window spans \(1250\) samples. 
We first concatenate the two modalities along the channel dimension, \(\mathbf{X} = [\mathbf{X}_{\mathrm{EMG}}; \mathbf{X}_{\mathrm{IMU}}] \in \mathbb{R}^{(C + D) \times N}\), and compute a short-time Fourier transform (STFT), which applies Fourier transforms over sliding temporal windows to obtain a time-frequency representation: $\mathbf{S} = \mathrm{STFT}(\mathbf{X}) \in \mathbb{R}^{(C + D) \times F \times T},$
where \(F\) is the number of frequency bins and \(T\) is the number of time frames. The spectrogram representation helps the network capture frequency features of muscle activity, and its 2-D image-like structure allows us to leverage pretrained vision encoders. As shown by the ablation in Appendix~\ref{app:ablation_model}, this spectrogram branch improves performance in our setup.

The time series input \(\mathbf{X}\) and spectrogram \(\mathbf{S}\) are then encoded separately. For the time series, we use a one-dimensional convolutional encoder, and for the spectrogram, we use a pretrained DINOv3 \cite{simeoni2025dinov3} encoder. The resulting features are concatenated and passed to a transformer decoder, which predicts the fingertip force \(F_{\text{ftp}} \in \mathbb{R}^{5 \times N}\) for all five fingers over the same temporal window. 
The estimated fingertip forces \(F_{\text{ftp}} \in \mathbb{R}^{5 \times N}\), paired with human video, are used to train robot policies.

\secv
\section{Force-Aware Policy Learning from Human Demonstrations}
\secv
\label{sec:policy_learn}

To convert force-enriched human demonstrations into robot-executable policies, we extend a flow matching transformer policy so that force enters as an additional channel on \emph{both} the observation side and the action side. 
As shown in Figure~\ref{fig:overview}, the pipeline takes synchronized human video and estimated fingertip forces as input, retargets human hand motion to a robot embodiment, and trains a flow-matching policy to predict both motion and force trajectories. This allows the robot to learn not only where to move but also how much force to apply during contact-rich manipulation.

\textbf{From Video to Embodied Observations.} 
We retarget per-frame human hand-object trajectories into an embodiment-agnostic representation that a parallel-jaw robot can execute. 
Hand keypoints define an $\mathrm{SE}(3)$ end-effector pose $T_{ee}$ and a 1-DoF aperture $g$, while each manipulated object is tracked and lifted to a 6-DoF pose. On the visual side, we segment and inpaint the human arm, then overlay the virtual gripper and tracked object keypoints, producing an RGB observation that hides human appearance while preserving scene layout. 
Pipeline and tool details are given in Appendix~\ref{app:tokens}.

\textbf{Flow Matching with Force-Augmented Action.}
The visual frame and a set of spatial-relation tokens jointly
condition a transformer decoder that predicts an action chunk over a
$K{=}50$-step horizon. We augment each action with grip force as an
additional prediction target: 
$a_t = [\,p\,;\,r_{6D}\,;\,g\,;\,f\,] \in \mathbb{R}^{11}$
where $p$ is the 3-D end-effector position, $r_{6D}$ is a continuous 6-D
rotation representation, $g$ is the grasp aperture, and $f$ is the desired grip force.
We train the decoder with conditional flow matching~\cite{lipman2023flow}:
it regresses a velocity field that transports a Gaussian prior to the
target action under per-component weights for pose, grasp,
and force. Because force appears in both the policy input and prediction target, the network learns a closed-loop relationship between the current contact state and future force commands, rather than treating force as a passive observation.

\secv
\section{Experiments}
\secv

\subsection{Ablation on Electrode Placement}
\secvv

\paragraphc{Baselines.} We compare our muscle-aware placement with representative wrist sEMG layouts: a single dorsal channel \cite{xiao2025wrist2finger}, two channels over opposing flexor and extensor regions \cite{tavakoli2018robust}, four wrist channels \cite{botros2022wrist}, and an evenly distributed eight-channel armband layout \cite{liu2021wrhand,pradhan2022grabmyo}.

\paragraphc{Protocol.}
For the 1-, 2-, and 4-channel settings, we use our released dataset and mask out unused channels. 
To compare even placement with our muscle-aware layout, since we cannot have both settings at the same time,
we collect two separate 30-minute datasets, one for each placement. 
Both use the same protocol: 10 objects, 2-, 3-, and 5-finger grasps, repeated 5 times per object. 

\paragraphc{Results.} As shown in Table~\ref{tab:force_error_newton}, increasing the number of channels consistently reduces force prediction error. Under the matched 30-minute protocol, our muscle-aware placement further improves over the evenly spaced 8-channel layout by 18\%. These results show that anatomically guided placement provides more informative EMG signals than uniform wrist coverage.

\begin{table}[t]
\centering
\caption{Average force prediction errors with different electrode placements.}
\vspace{0.2em}
\label{tab:force_error_newton}
\setlength{\tabcolsep}{4pt}
\begin{minipage}[t]{0.32\linewidth}
\centering
\resizebox{\linewidth}{!}{%
\begin{tabular}{lcc}
\toprule
Hardware & MAE (N) & RMSE (N) \\
\midrule
1 ch      & 1.89 & 3.69 \\
2 ch      & 1.76 & 3.06 \\
4 ch      & 0.93 & 2.00 \\
8 ch ours & 0.85 & 1.92 \\
\bottomrule
\end{tabular}}
\end{minipage}
\hspace{1em}
\begin{minipage}[t]{0.32\linewidth}
\centering
\resizebox{\linewidth}{!}{%
\begin{tabular}{lcc}
\toprule
Hardware  & MAE (N) & RMSE (N) \\
\midrule
8 ch even & 0.94 & 1.77 \\
8 ch ours & 0.77 & 1.33 \\
\bottomrule
\end{tabular}}
\end{minipage}

\vspace{0.3em}
\footnotesize \textit{Note:} The left and right sub-tables should \textbf{not} be directly compared as they are using different data.
\vspace{-0.2cm}
\end{table}

\secvv
\subsection{Contact and Force Estimation}
\secvv

\begin{figure*}[ht]
	\centering
	\includegraphics[width=\textwidth]{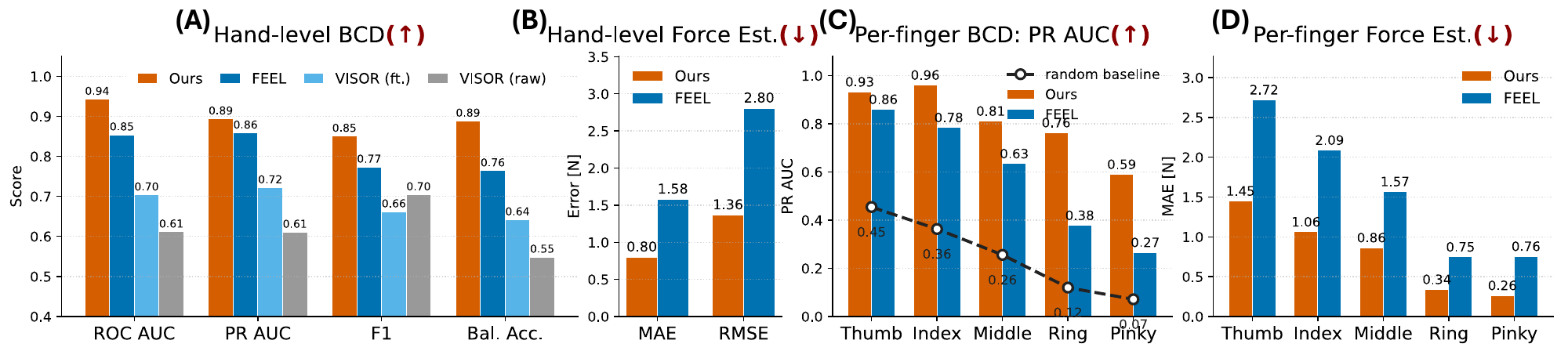}
	\caption{
		Quantitative force estimation results.
	}
	\label{fig:force}
	\vspace{-0.5cm}
\end{figure*}

\paragraphc{Baselines.} We compare against two vision-based estimators.
VISOR-HOS~\cite{darkhalil2022epic} is pre-trained on
EPIC-KITCHENS; we evaluate both the released checkpoint (raw) and a version
fine-tuned on our data (ft.). FEEL~\cite{dessalene2026feel} trains a 2-layer
MLP on frozen DINOv3~\cite{simeoni2025dinov3} features for binary contact detection (BCD)
and per-finger force regression. The contact threshold is $0.8$\,N
for all binary metrics.

\paragraphc{Results} (Fig.~\ref{fig:force}).
\AlgName{} outperforms every vision baseline on hand-level BCD, and on
hand-level force regression sEMG roughly halves both error metrics, as shown in \textbf{hand-level results}
(Fig.~\ref{fig:force} A,B). The advantage compounds at \textbf{finger level}
(Fig.~\ref{fig:force} C,D), where sEMG recovers load on muscles a camera
cannot see. We report PR AUC, the area under the precision-recall curve, to summarize contact-detection quality under class imbalance. On the ring ($0.763$ vs.\ $0.398$) and pinky ($0.590$
vs.\ $0.314$), \AlgName{} exceeds FEEL by more than $2\times$ and the prevalence-matched
random baseline by over $6\times$, with consistent gains across all five
fingers. PR AUC is the more informative metric in this imbalanced setting: positive prevalence
falls from $45\%$ (thumb) to $7\%$ (pinky), and ROC AUC compresses above
$0.85$, masking precisely the regime in which vision breaks down.
Overall, vision provides useful contact cues, but it cannot reliably infer hidden force magnitudes. The wrist-worn sEMG band complements vision by capturing muscle activity directly, producing per-finger force traces suitable for downstream force-aware policy learning.

\secvv
\subsection{Force-aware Policy Learning}
\secvv

\begin{table}[t]
\centering
\small
\setlength{\tabcolsep}{3.2pt}
\renewcommand{\arraystretch}{1.12}
\caption{\textbf{Policy success on pick, squeeze, and place tasks.}}
\vspace{0.3em}
\label{tab:policy_learning}
\resizebox{\linewidth}{!}{%
\begin{tabular}{lccccccccc}
\toprule
\textbf{Object} 
& \textbf{Mustard} 
& \textbf{\makecell{Face\\Wash}} 
& \textbf{Toothpaste} 
& \textbf{\makecell{BBQ\\Small}} 
& \textbf{Chips} 
& \textbf{BBQ} 
& \textbf{\makecell{Mustard\\Small}} 
& \textbf{Chocolate} 
& \textbf{Ketchup} \\
\midrule
Weight (g) 
& \textbf{650} & 280 & 176 & 290 & \textbf{43} & 580 & 244 & 210 & 410 \\
Grasp Width (mm) 
& 49 $\sim$ 54 & 18 $\sim$ 58 & 9 $\sim$ 37 & 54 $\sim$ 56 & \textbf{1} $\sim$ 68 & 67 $\sim$ \textbf{72} & 37 $\sim$ 40 & 19 $\sim$ 22 & 44 $\sim$ 46 \\
\midrule
Continuous gripper 
& 7-0 & 6-4 & 4-2 & 6-0 & 3-N/A & 8-0 & 3-0 & 0-0 & 7-0 \\
Binary gripper     
& 10-0 & 10-0 & 8-0 & 9-0 & 10-N/A & 7-0 & 10-0 & 10-0 & 10-0 \\
\textbf{\AlgName{}}
& \textbf{10-10} & \textbf{10-10} & \textbf{9-9} & \textbf{8-8} & \textbf{7-N/A} 
& \textbf{6-6} & \textbf{10-10} & \textbf{10-9} & \textbf{10-10} \\
\bottomrule
\end{tabular}%
}
\vspace{2pt}
\begin{minipage}{0.98\linewidth}
\vspace{0.4em}
\footnotesize \textit{Note:} Each entry reports \textbf{pick-and-place} and \textbf{squeeze} success out of 10 trials. \textbf{N/A} indicates that squeezing is not required for the task. \textbf{Grasp Width} denotes the gripper aperture range from initial contact to 20N of applied force.
\end{minipage}
\vspace{-0.6cm}
\end{table}

\paragraphc{3-Step Deployment and Experimental Setup. }
After the hardware and pretrained EMG2Force model is prepared, deployment is simple: the user performs a 15-minute calibration with \AlgName{} and fingertip force data, then collects target-task demonstrations using only \AlgName{} and video, while EMG2Force labels these demonstrations with per-finger force traces for force-aware policy training. At robot deployment, the learned policy predicts both motion and force trajectories. 
More details are in Appendix \ref{app:depoly}. 
We evaluate \AlgName{} on a UR-5 robot, operating in a real-world tabletop manipulation environment. One ZED 2i camera is mounted to provide third-person RGB-D observations. We collect 15 human demonstrations per object and train the flow-matching policy of Section \ref{sec:policy_learn}. 
Details of the setup and controllers are provided in Appendix \ref{app:exp_setup_controller}.

\textbf{Baselines.} Similar to \cite{feeltheforce}, we compare \AlgName{} with two gripper-control baselines that use the same visual observations and retargeted motion representation. \textbf{Binary Gripper} predicts a discrete open or close command. \textbf{Continuous Gripper} predicts a continuous gripper aperture, testing whether gripper position can serve as an implicit proxy for contact force. 

\textbf{Task Description.} We evaluate on a pick-squeeze-place task, which requires time-varying, object-specific force control. In each rollout, the robot must pick up an object, apply the appropriate squeeze force, and place it at the target location. Since the three stages require different force regimes, we define success at two levels: pick and place success requires the robot to stably grasp, transport, and place the object at the target location without dropping it, while squeeze success requires the robot to apply an object-specific force profile that human evaluators can distinguish from the forces used for picking and placing, indicating a purposeful squeeze rather than a fixed grasp command.
The task cannot be solved by kinematics alone: insufficient force causes slipping or incomplete squeezing, while excessive force can deform the object, destabilize the grasp, or produce incorrect interaction behavior. We test nine objects with diverse shapes, sizes, and weights, including both in-distribution objects and out-of-distribution (OOD) objects. Since the desired force varies across objects, this benchmark directly tests whether the policy can use force supervision from human data rather than relying on a fixed grasp command.

\paragraphc{\AlgName{} applies object-specific forces that gripper-only baselines cannot reproduce.} 
As shown in Table~\ref{tab:policy_learning}, \AlgName{} succeeds across objects with diverse weight, grasp width, and compliance. The binary gripper baseline can often pick and place objects, but cannot produce the required squeeze behavior. The continuous gripper baseline can sometimes squeeze soft objects, where deformation makes aperture changes partially correlate with force, but it fails on rigid objects and remains unreliable overall. This is partly because gripper position estimated from human video is noisy under finger occlusion. In contrast, \AlgName{} directly predicts force from wrist sEMG, enabling more reliable object-specific force control.

\begin{figure*}[t]
	\centering
	\includegraphics[width=\textwidth]{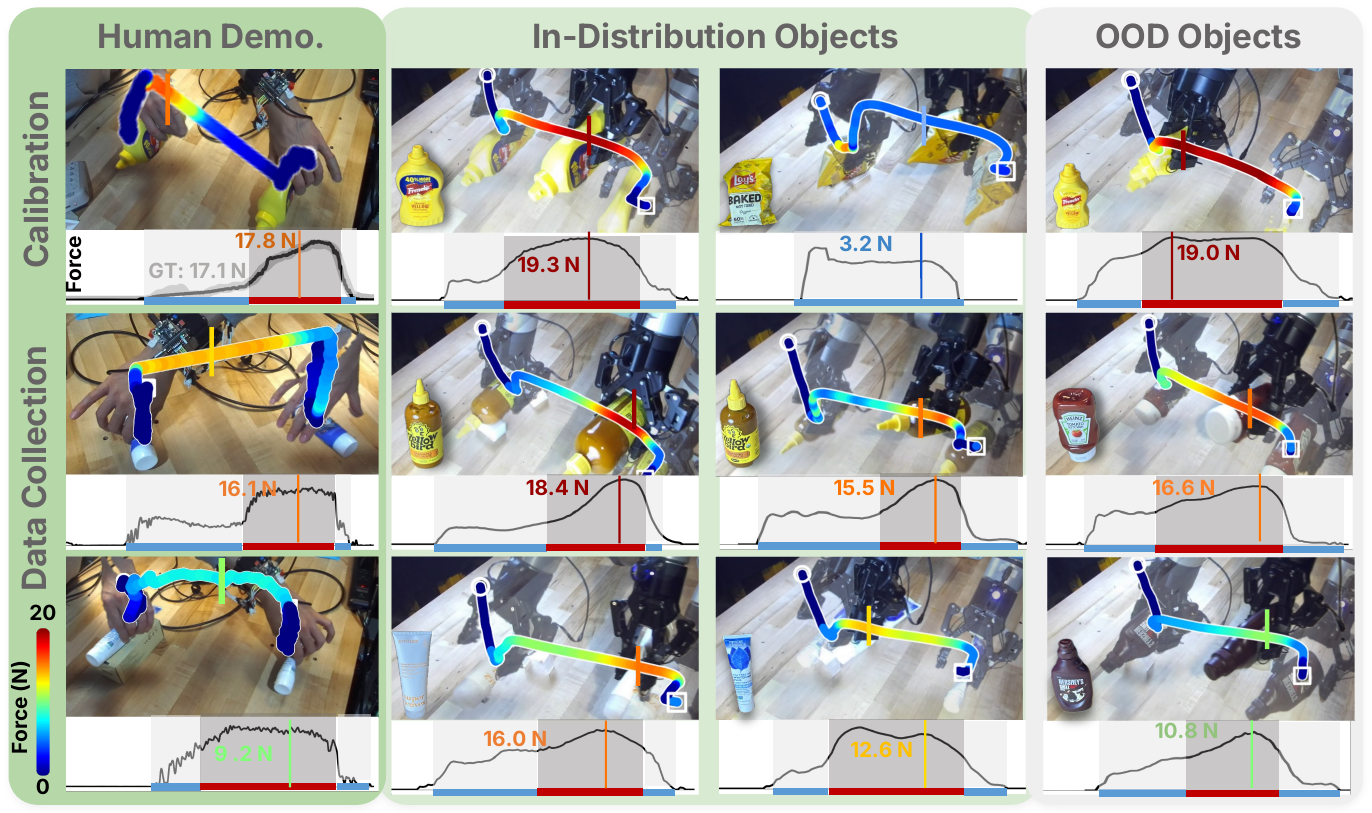}
    \vspace{-0.6cm}
    \caption{\textbf{Force-aware robot policy rollouts.}
    \AlgName{} predicts object-specific force for pick, squeeze, and place tasks across objects with grasp widths from 1 to 72~mm, including both ID and OOD objects. In each force plot, different background colors indicate different task stages: pick, squeeze, and place. Across objects, the policy produces distinct peak forces from 3.2~N to 19.3~N, enabling squeeze behaviors that baselines fail to reproduce.}
	\label{fig:roboexp}
	\vspace{-0.72cm}
\end{figure*}

\paragraphc{Predicted force are object-specific and transfer to OOD objects.} Figure~\ref{fig:roboexp} visualizes rollouts on both in-distribution and OOD objects. Across objects, \AlgName{} predicts distinct force profiles instead of collapsing to a constant grasp command, with peak forces ranging from light contacts around 3.2~N to stronger squeezes around 19~N. The OOD examples show that the learned policy can still produce meaningful force trajectories for objects not seen during policy training, suggesting that the force channel provides a transferable interaction representation beyond visual appearance or object identity. Overall, these results demonstrate that wrist sEMG can turn natural human demonstrations into force-aware supervision for robot learning, enabling forceful manipulation behaviors that motion-only or gripper-only policies cannot achieve.

\secv
\section{Related Work}
\secv

\paragraphc{Contact and Interaction Understanding.}
Existing solutions can be broadly categorized into two approaches: vision-based estimation and tactile gloves. 
For vision-based methods, some predict binary contact from monocular video \citep{narasimhaswamy2020detecting,yagi2021hand}, while others estimate richer 3D contact maps or hand-object interaction using annotated datasets \citep{hampali2020honnotate,brahmbhatt2020contactpose} or generate pseudo-contact labels \citep{tse2022s2contact}, and recent work further explores vision to touch prediction from egocentric or wrist mounted video \citep{touchanything}.
Other works incorporate direct tactile sensing through wearable systems such as OpenTouch and OSMO, which combine egocentric video with full-hand tactile sensing for contact-rich manipulation and human-to-robot transfer \cite{opentouch,osmo}. 
However, vision-based force estimation remains ambiguous under occlusion and motion blur, while tactile gloves require direct instrumentation of the hand.

\paragraphc{sEMG for Force Estimation.}
sEMG has been widely explored for gesture recognition \cite{emg2pose, zhao2026dexemg, xu2024chatemg, wang2025reactemg, verma2025emg2tendon, yang2025high} and grasp detection \cite{wang2025chainofmodality}, but many prior works focus on binary actions such as hand open/close rather than fine-grained force estimation. Existing systems often rely on single-channel EMG or uniformly distributed electrodes \cite{sun2018one, xiao2025wrist2finger}, limiting spatial selectivity and robustness to placement shifts, while other approaches use task-specific signal-processing pipelines that generalize poorly across manipulation scenarios. Recent work shows that muscle-aware electrode placement and multimodal sensing improve force estimation quality: targeted channel selection based on forearm anatomy outperforms circumferential layouts for hand and finger control \cite{pelaez2022reducing, cho2022training}, and combining sEMG with inertial sensing helps disambiguate muscle activations during dynamic manipulation \cite{xiao2025wrist2finger, mao2021simultaneous}.

\paragraphc{Robot Learning From Human Videos.}
Prior work reduces the embodiment gap through visual transfer, cross-embodiment alignment, and embodiment-agnostic representations. Early methods translate human videos into robot-compatible visual domains \citep{lbw,hudor}, while later approaches align observations or actions across humans and robots \citep{whirl}. More recent methods represent manipulation with keypoint tracks and interaction-centric features \citep{kruger2011object, pointpolicy,p3po,hop,humanego,aina}. 
These representations enable scalable policy learning from human videos, but they remain primarily visual or kinematic and do not directly capture the contact forces that are critical for forceful manipulation. In parallel, force-aware policy learning has been explored with fingertip force sensing gloves \citep{feeltheforce}; however, such fingertip instrumentation requires specialized contact sensors and can alter natural hand-object interaction, limiting scalability compared with wrist-worn sensing approaches such as ours.

\secv
\section{Conclusion}
\label{sec:conclusion}
\secv

We presented \AlgName{}, a low-cost wrist-worn sEMG system that turns natural
human manipulation into force-enriched demonstrations for robot learning.
By combining muscle-aware electrode placement, an IMU, and a
spectrogram-augmented EMG2Force model, \AlgName{} recovers per-finger contact
forces that vision cannot reliably infer, especially under occlusion. After a
short calibration, demonstrations are collected with only \AlgName{} and video,
and the force-enriched demonstrations are retargeted to a robot embodiment to train a force-aware policy. Muscle-aware placement reduces force error by $18\%$
over an evenly spaced eight-channel layout, EMG2Force more than halves the
error of vision-based baselines, and the resulting policy achieves
object-specific squeeze control that motion-only and gripper-only baselines
cannot reproduce, transferring even to OOD objects. Overall,
wrist sEMG offers a scalable and unobtrusive path to adding force supervision
to human demonstration data for forceful robot manipulation.

\secvv
\subsection*{Limitations and Discussion}
\secvv

\textbf{Absolute force accuracy.}
Force prediction from sEMG is not as accurate as direct force sensing, but its current accuracy is sufficient for many forceful manipulation tasks. More importantly, the force prior learned from human data helps the policy understand when and how hard to apply force, and residual errors can be further reduced with limited robot interaction data or reinforcement learning using the robot's own force feedback.

\textbf{Cross-user generalization.} 
Muscle signals vary across users, so our current pipeline involves 4 users and applies per user calibration. This limitation may be mitigated by scaling data collection: large-scale sEMG datasets have shown strong cross-user generalization~\cite{emg2pose}, and related wrist sEMG controllers now operate across users without calibration~\cite{metaneuralband}. We therefore expect larger EMG2Force datasets to reduce or remove the need for user-specific calibration.

\textbf{Calibration still requires fingertip force sensors.}
Calibration currently relies on fingertip force sensors for ground truth, so setup is not yet fully sensor-free. This requirement may be reduced in two ways: by scaling EMG2Force to a larger and more diverse user population, or by developing a sensor-free calibration protocol based on standardized grasp interactions with calibrated objects or fixtures. The former follows the same scaling argument above: a sufficiently general model may reduce or remove the need for per-user calibration. The latter is analogous to the guided calibration poses used by commercial hand-tracking gloves~\cite{manus}.

\acknowledgments{We thank Jiaxi Zheng, Eadom Dessalene, and Levi Burner for their helpful support and feedback throughout this project.}

\bibliography{references}

\clearpage
\appendix
\section*{Appendix}

\secv
\section{Electrode Placement Details}
\secv
\label{app:placement} 

\begin{figure*}[ht]
	\centering
	\includegraphics[width=0.7\textwidth]{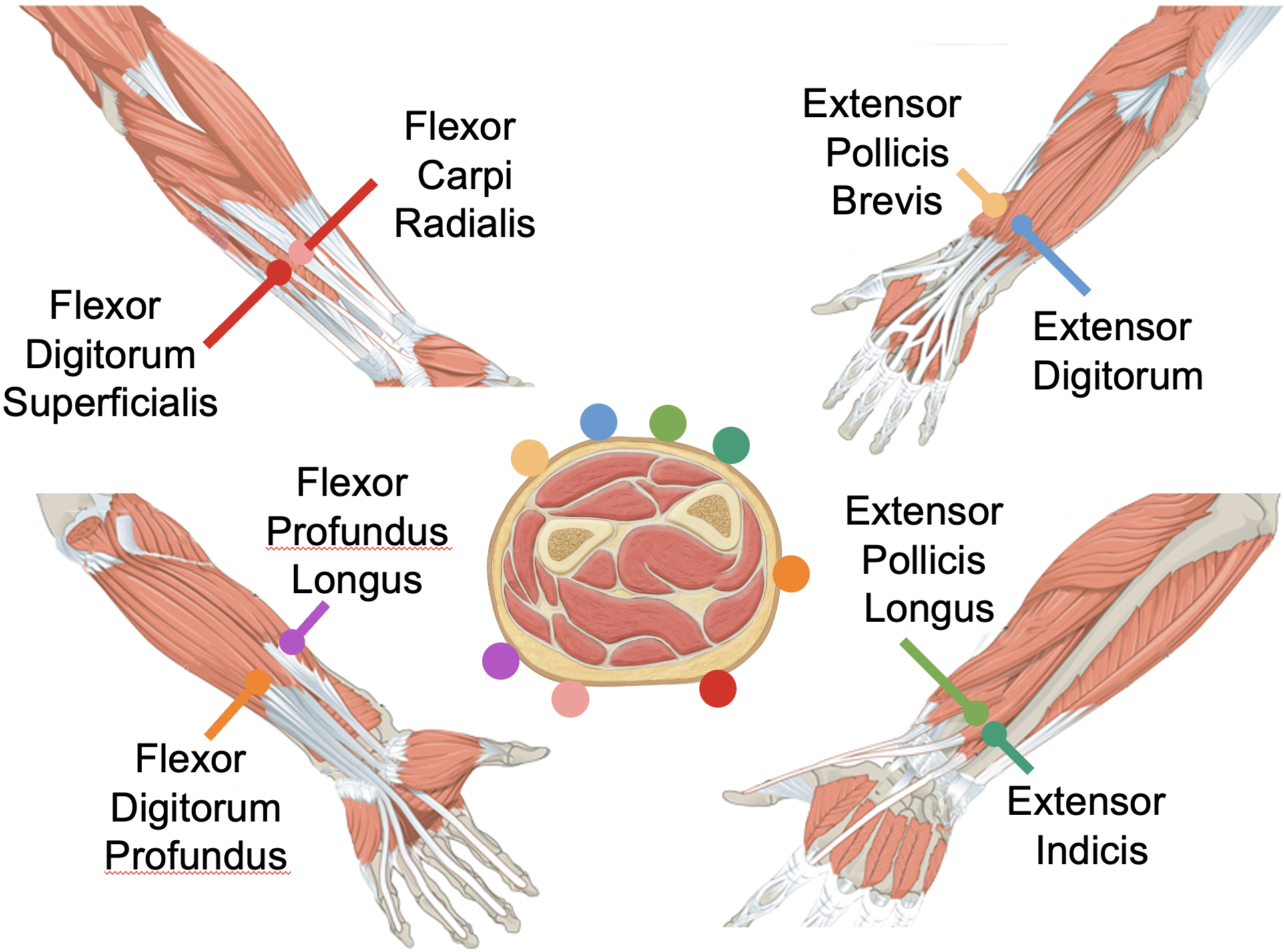}
	\caption{
		Electrode placement details.
	}
	\label{fig:app_placement}
	\vspace{-0.5cm}
\end{figure*}

We use eight sEMG channels: seven to capture muscle activity associated with fingertip control and one to capture wrist flexion (Figure \ref{app:placement}). Channel 1 is placed over the extensor pollicis brevis (EPB) for thumb metacarpophalangeal (MCP) extension; Channel 2 over the extensor digitorum (ED) for MCP extension of the index, middle, ring, and little fingers; Channel 3 over the extensor pollicis longus (EPL) for thumb interphalangeal (IP) extension; Channel 4 over the extensor indicis for index-finger extension; Channel 5 over the flexor digitorum profundus (FDP) for flexion at the distal interphalangeal (DIP) joints of the index, middle, ring, and little fingers; Channel 6 over the flexor digitorum superficialis (FDS) for flexion at the proximal interphalangeal (PIP) joints of the same fingers; Channel 7 over the flexor carpi radialis (FCR) for wrist flexion; and Channel 8 over the flexor pollicis longus (FPL) for thumb IP flexion. A reference electrode over the ulnar styloid process serves as the common ground. Before placement, the skin is cleaned with an alcohol wipe to remove oils and reduce skin–electrode impedance. After donning, a series of hand gestures (Fig. 3) validates correct sensor positioning on the intended muscle sites, safeguarding the accuracy and consistency of subsequent measurements.

\secv
\section{Hardware Extensibility}
\label{app:hardware_extensibility}
\secv

\begin{figure}[t]
\centering
\includegraphics[width=\linewidth]{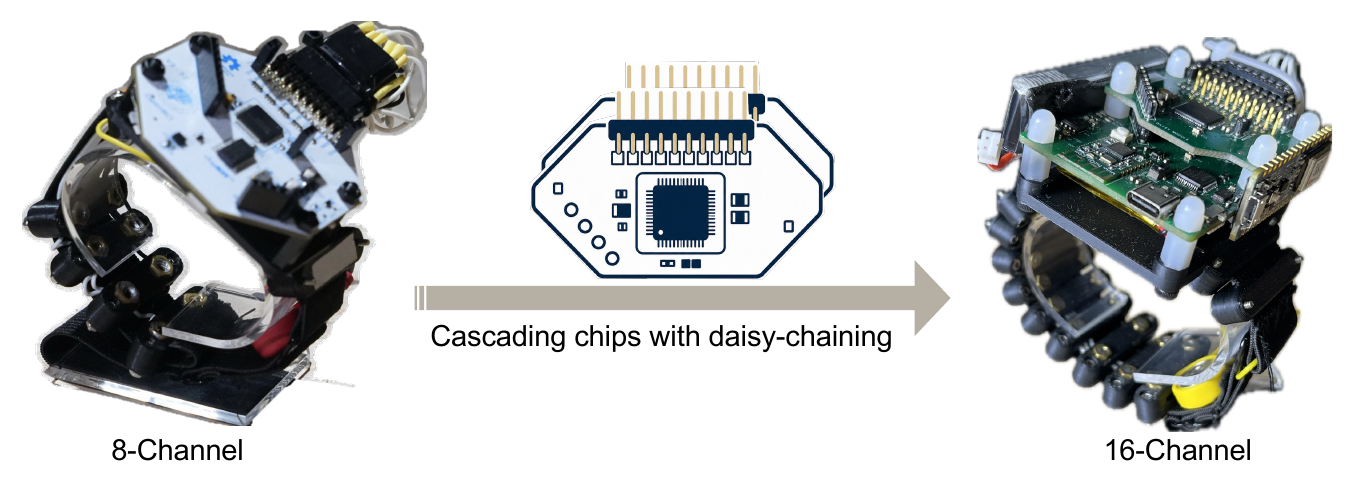}
\caption{\textbf{Hardware extensibility through daisy chaining.}
The 8-channel acquisition configuration used in this work can be extended by cascading a second acquisition module through a daisy-chain interface, forming a 16-channel configuration. This modular design supports denser and task-specific electrode layouts without redesigning the overall wearable platform.}
\label{fig:app_extend}
\vspace{-0.4cm}
\end{figure}

The 8-channel configuration used in our experiments was selected as a practical balance between muscle coverage, wearability, and system complexity. However, the wearable acquisition platform is not limited to a fixed channel count. As shown in Figure~\ref{fig:app_extend}, the electronics support modular expansion through daisy chaining: an additional 8-channel acquisition module can be cascaded with the primary board to form a 16-channel configuration.

This extensibility enables denser forearm coverage, alternative anatomically guided electrode layouts, or additional sensing sites for task-specific muscle groups when greater spatial selectivity is needed. In an expanded configuration, EMG2Force can be trained and calibrated using the corresponding larger channel set, while the remaining demonstration collection, force labeling, and robot policy-learning pipeline remains unchanged. Thus, \AlgName{} is designed as a scalable sensing platform rather than a system tied to the particular 8-channel layout evaluated in this work.

\secv
\section{Details of Spatial-Relation Tokens and Embodiment Transfer}
\secv
\label{app:tokens}

\paragraphc{From video to embodied observations.}
Per-frame 3-D hand keypoints from Aria MPS~\cite{engel2023aria} are
retargeted to a parallel-jaw end-effector. The midpoint of the thumb
and index fingertips defines the gripper position; the
metacarpophalangeal (MCP) joints---which remain well-separated through
the full pinch cycle---define a Gram--Schmidt orientation frame; and the
normalized thumb--index distance gives a 1-DoF aperture. Each frame thus
produces an $\mathrm{SE}(3)$ end-effector pose $T_{ee}$ and a grasp
scalar $g$ that match what a parallel-jaw robot can execute. Each tracked
object is detected with Grounding DINO~\cite{liu2024groundingdino},
segmented with SAM2~\cite{ravi2024sam2}, tracked across frames with
CoTracker~\cite{karaev2024cotracker}, and lifted to a 6-DoF pose using
Orient-Anything~\cite{wu2025orientanything}. On the visual side, we segment the
human arm with SAM2 and inpaint it with LaMa~\cite{suvorov2022lama}, then
overlay the virtual gripper together with the tracked object keypoints
into the inpainted image, producing an embodiment-agnostic RGB
observation that hides the human appearance while preserving scene
layout.

\paragraphc{Spatial-relation tokens with force fusion.}
We encode the scene as a set of per-entity tokens, one for each hand and
each tracked object. For entity $k$ the token concatenates (i) an
entity-type indicator, (ii) the entity's 6-DoF pose in a shared scene
frame, (iii) the hand pose expressed in the entity's own local
frame---so the sub-vector evolves with the manipulation phase as the hand
approaches, contacts, or transports the entity---and (iv) a per-entity
scalar (grasp state for hands, a sentinel for objects). This construction
makes every token explicitly carry the spatial relation between an entity
and the manipulator rather than treating them as independent points. We
attach an extra force channel to the hand token only: the 1-D grip force
$f_t$, taken as the average of the thumb and index forces in
$F_{\text{ftp}}$ and time-aligned with the visual frame. Force is bound
to the hand because it is physically a quantity the hand exerts on
objects; placing it on the same token that carries the gripper pose lets
the policy condition every action on the current contact state.

\secv
\section{Ablation Study on EMG2Force Model}
\secv
\label{app:ablation_model}

We ablate the main input and representation choices in EMG2Force by comparing variants with or without the spectrogram branch and with or without IMU input. The spectrogram branch tests whether frequency-domain muscle activity provides complementary information beyond raw time-domain sEMG signals, while the IMU ablation tests whether motion cues help disambiguate muscle activation during dynamic manipulation. All variants use the same training and evaluation protocol, and only the corresponding input branch or modality is removed.

\begin{figure*}[h]
	\centering
	\includegraphics[width=\textwidth]{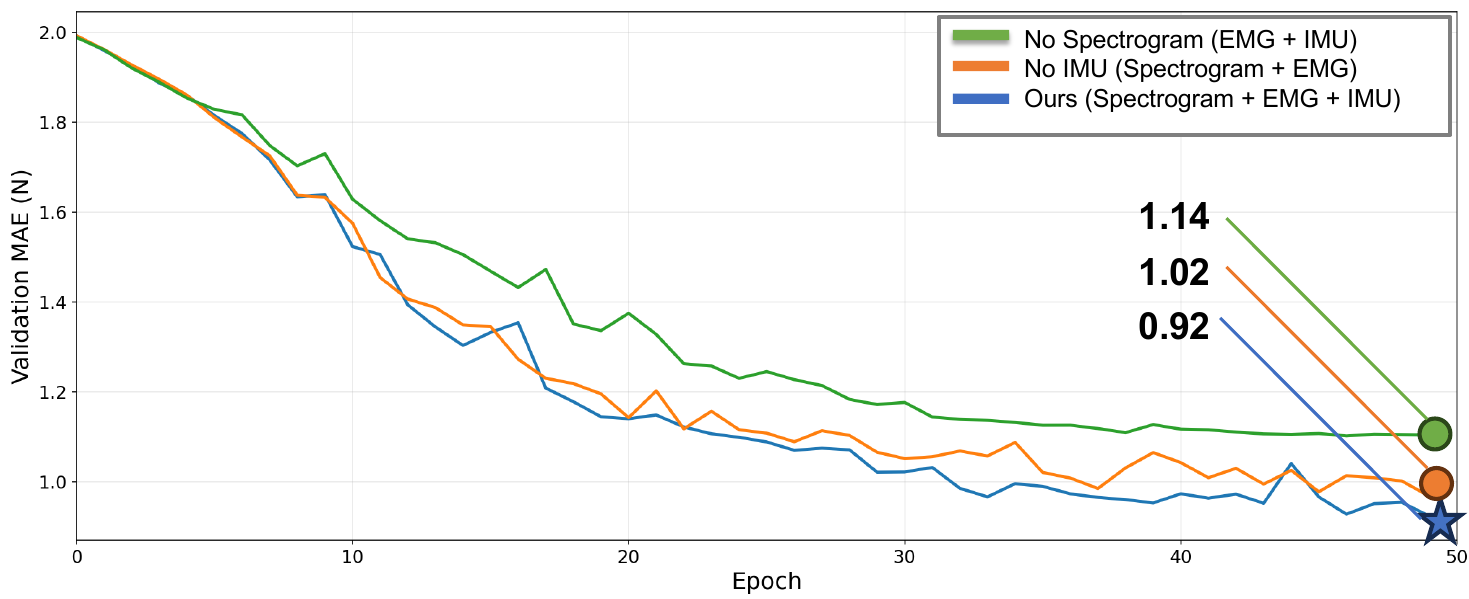}
	\caption{
		We compare the full model against variants that remove either the spectrogram branch or IMU input. The full model achieves the lowest MAE, while removing the spectrogram branch or IMU input increases force prediction error. This shows that frequency-domain sEMG features and motion cues provide complementary information for fingertip force estimation.
	}
	\label{fig:app_ablation}
\end{figure*}

As shown in Figure~\ref{fig:app_ablation}, the full model, which combines time-domain sEMG, spectrogram features, and IMU signals, achieves the lowest force prediction error with an MAE of 0.92~N. Removing the spectrogram branch increases the error to 1.14~N, suggesting that frequency-domain patterns provide useful information for estimating fingertip force. Removing IMU input increases the error to 1.02~N, indicating that wrist motion cues are important for separating force-related muscle activity from motion-induced signal changes. These results support our design choice of using a multimodal time-frequency representation for wrist-based force estimation.

\secv
\section{Details of the Three-Step Deployment}
\secv
\label{app:depoly}

\begin{figure*}[ht]
	\centering
	\includegraphics[width=\textwidth]{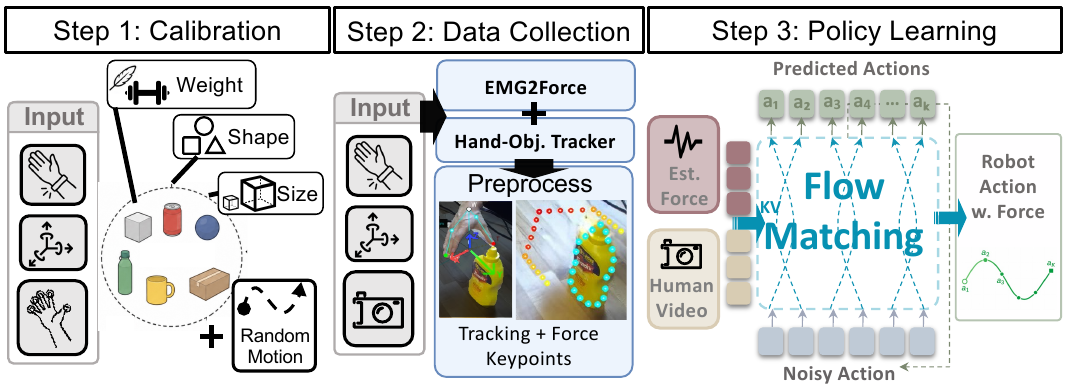}
	\caption{
		Details about the three-step deployment.
	}
	\label{fig:app_deploy}
\end{figure*}

After the hardware and pretrained EMG2Force model is prepared, deployment only requires three steps: a short user-specific calibration, data collection and policy training, as shown in Figure~\ref{fig:app_deploy}.

\textbf{Step 1. }
The user wears the \AlgName{} and collects about 15 minutes of paired sEMG, IMU, and fingertip force data. This calibration adapts the force estimator to the user’s muscle activation patterns and sensor placement.

\textbf{Step 2. }
Once calibrated, the fingertip force sensors are no longer needed. The user can collect daily manipulation demonstrations with only \AlgName{} and video. The calibrated model converts these signals into per-finger force traces, producing force-augmented human demonstrations. These demonstrations are then mapped into the unified human-to-robot representation and used to train the force-aware robot policy.

\textbf{Step 3. }
During robot deployment, the policy predicts both motion and force conditioned on visual observations, while the robot gripper force sensor provides force feedback for closed-loop execution. In this way, once the hardware is available, a user only needs a 15-minute calibration before conveniently collecting force-enriched human demonstrations and training robot policies.
\secv
\section{Details of the Experimental Setup and Controllers}
\secv
\label{app:exp_setup_controller}

\begin{figure*}[h]
	\centering
	\includegraphics[width=\textwidth]{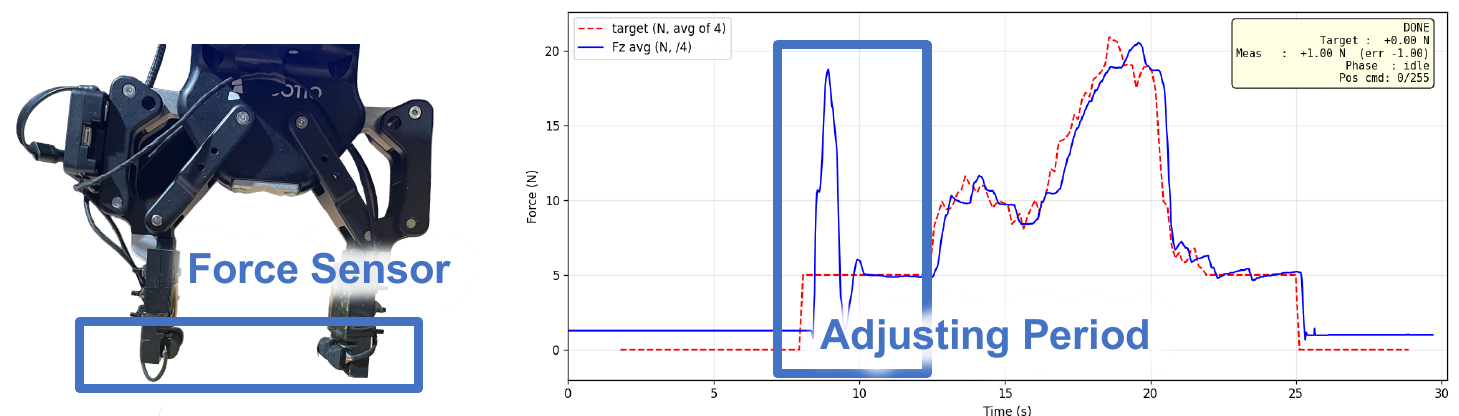}
    \caption{\textbf{Robot fingertip force sensing and force-control procedure.}
    Left: because the Robotiq gripper does not provide sufficiently precise and timely fingertip force feedback, we attach four Paxini force sensors to the gripper fingertips. Right: when the policy predicts a close command, we pause execution for a short adjustment period and regulate the gripper to a 5~N pre-grasp force. After this stable contact is established, the policy resumes and a PD controller tracks the predicted force trajectory during the squeeze and place stages.}
	\label{fig:app_force_track}
\end{figure*}

We evaluate \AlgName{} in a real-world tabletop manipulation setup with a UR-5 robot and a Robotiq parallel-jaw gripper. A ZED 2i camera is mounted in front of the workspace to provide third-person RGB-D observations. As described in the main text, we collect 15 human demonstrations per object and train the flow-matching policy to predict both motion and force trajectories.

The object set is listed in Table~\ref{tab:policy_learning}. It includes nine everyday objects with diverse weights, shapes, sizes, materials, and grasp widths: mustard, face wash, toothpaste, small BBQ sauce, chips, BBQ sauce, small mustard, chocolate, and ketchup. The objects span light and heavy items, rigid and deformable containers, and narrow to wide grasp widths. This diversity makes the task sensitive to object-specific force control: the robot must apply enough force to avoid slipping, but avoid collapsing soft objects or applying an incorrect squeeze profile.

\paragraphc{Force sensing on the robot gripper.}
Although the Robotiq gripper provides reliable position control, it does not provide precise fingertip force sensing and its force response is not timely enough for accurate force tracking. We therefore attach four Paxini force sensors to the gripper fingertips, as shown in Figure~\ref{fig:app_force_track}. These sensors provide direct contact-force feedback during robot execution and allow the robot to track the force command predicted by the policy.

\paragraphc{Pre-grasp adjustment and force tracking.}
Directly tracking force from the beginning of a rollout is unstable because the gripper may not yet be in firm contact with the object, and the Robotiq gripper has delayed force response. To address this, whenever the policy predicts a close command, we temporarily pause the policy and enter a short adjustment period. During this period, the gripper closes around the object and adjusts its contact force to a fixed pre-grasp force of 5~N. After the object is stably grasped, the policy resumes execution. During the subsequent squeeze and place stages, we use a PD force controller to track the target force predicted by the policy, adjusting the gripper aperture according to the force error. As shown in Figure~\ref{fig:app_force_track}, this pre-grasp adjustment stabilizes the initial contact and allows the measured force to follow the desired force trajectory more closely during execution.

\secv
\section{Qualitative EMG2Force Predictions on the Pretraining Dataset}
\secv

\begin{figure*}[t]
	\centering
	\includegraphics[width=\textwidth]{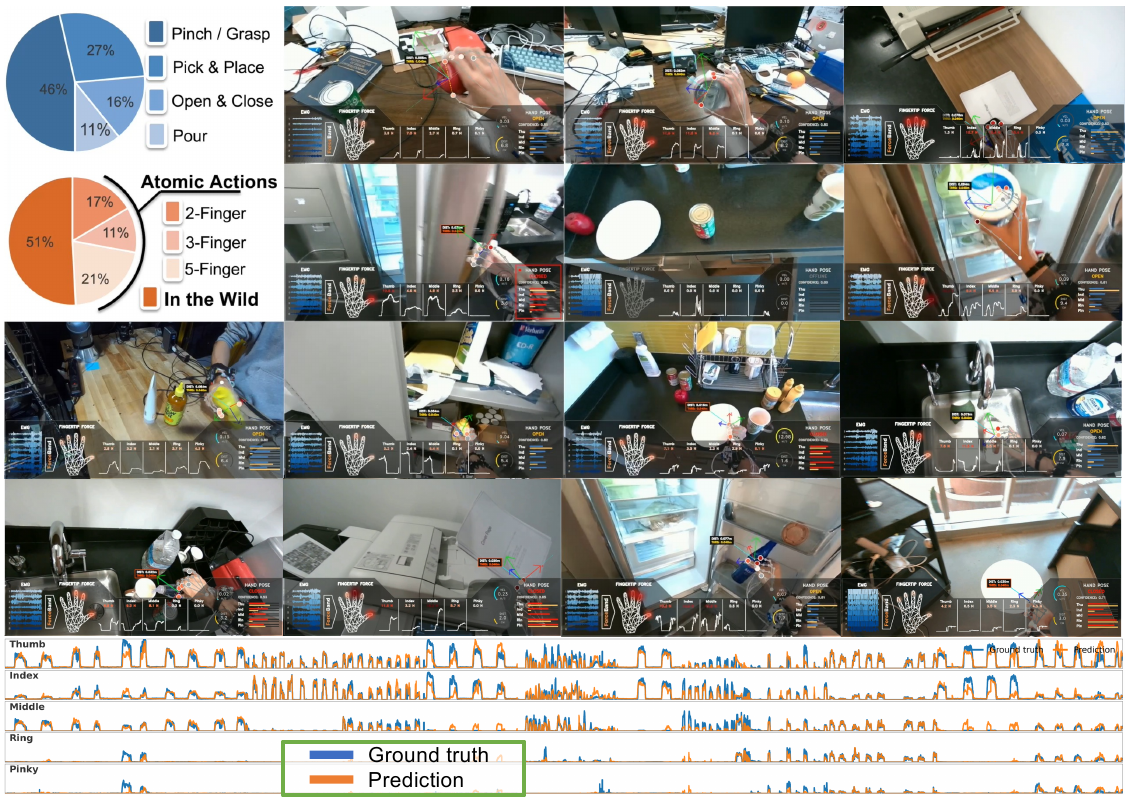}
    \caption{\textbf{Qualitative EMG2Force predictions on the pretraining dataset.}
    We show representative examples from the multimodal pretraining dataset, covering diverse action categories, gesture types, objects, and in-the-wild interactions. \textbf{All force curves overlaid on the images are EMG2Force estimates from \AlgName{} signals}, rather than direct fingertip force-sensor measurements. The bottom row shows a representative comparison between estimated force and ground-truth fingertip force, illustrating that the model captures the main timing and magnitude trends of contact, squeezing, and release across fingers.}
	\label{fig:app_pretrain}
    \vspace{-1.5em}
\end{figure*}

We provide additional qualitative results of EMG2Force on the pretraining dataset in Figure~\ref{fig:app_pretrain}. As described in Section~3, the dataset contains synchronized egocentric video, sEMG, IMU, and per-finger force measurements across diverse actions, gesture types, objects, and free-form daily interactions. These examples therefore test whether the model can recover force patterns beyond isolated canonical grasps.

\textbf{In these qualitative frames, all force curves overlaid on the images are estimated by EMG2Force from \AlgName{} signals}, rather than directly measured by fingertip force sensors. This visualization shows how the model can label natural demonstrations with per-finger force traces after calibration, after fingertip sensors are removed. Across different manipulation settings, the estimated forces follow the expected task structure: force increases during contact or squeezing, decreases after release, and remains low during non-contact intervals. The predictions also capture finger-specific activation patterns, such as stronger thumb and index forces during pinch-like interactions and broader multi-finger activation during grasping or object handling.

To better illustrate force-estimation accuracy, the bottom row additionally shows a representative clip comparing EMG2Force-estimated forces with synchronized ground-truth fingertip forces. Although some small peaks are smoothed or slightly shifted in time, especially for weaker ring and pinky contacts, the predictions preserve the main force events and relative finger usage. These qualitative results support the quantitative findings in the main text, showing that EMG2Force learns meaningful per-finger force dynamics from wrist sEMG and IMU signals across diverse pretraining demonstrations.

\section{Generalization Test}
\begin{figure*}[t]
	\centering
	\includegraphics[width=\textwidth]{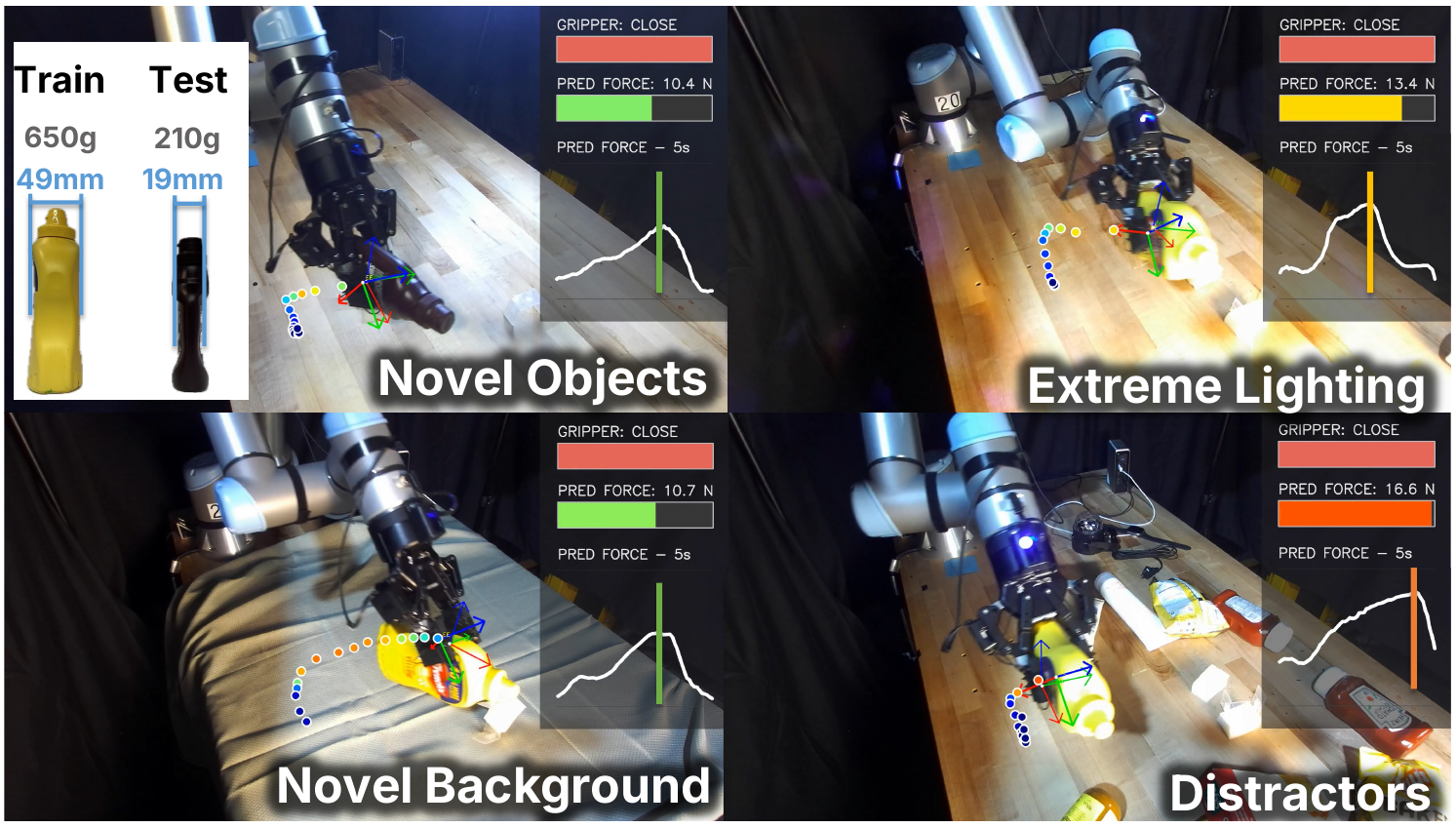}
    \caption{\textbf{Generalization test under visual and object-level distribution shifts.}
    We evaluate the learned policy on novel backgrounds, novel objects, extreme lighting, and visual distractors. In all cases, the robot follows the correct pick-squeeze-place trajectory and preserves the three-stage task structure. Background and texture changes can still affect the precise force magnitude, suggesting that visual appearance contributes to force prediction.}
	\label{fig:app_gen}
    \vspace{-1em}
\end{figure*}

We further evaluate whether the learned force-aware policy remains robust under visual and object-level distribution shifts. As shown in Figure~\ref{fig:app_gen}, we test four challenging settings: a novel background, novel objects, extreme lighting, and visual distractors. Across all cases, the policy executes the correct overall trajectory and preserves the three-stage task structure: picking up the object, applying a squeeze, and placing it at the target location. This suggests that the learned policy does not simply memorize the training scene but can generalize the high-level manipulation behavior to new visual conditions.

However, we also observe that visual changes, especially background and texture shifts, can affect the exact force applied during the squeeze stage. This indicates that the policy still uses image texture as part of its force-conditioned decision process. Such sensitivity is expected because our model intentionally leverages rich visual observations rather than discarding appearance information. This creates a trade-off: using dense image texture improves object and scene understanding, but can also introduce some force variation under visual distribution shifts. Overall, the policy generalizes the task structure reliably, while more robust force calibration under visual domain shifts remains an important direction for future work.

\end{document}